\documentclass[runningheads]{llncs}
\usepackage[T1]{fontenc}
\usepackage{graphicx}
\usepackage{graphicx}
\usepackage{wrapfig}
\usepackage{float} 
\usepackage{booktabs}
\usepackage{array}
\usepackage{geometry}
\usepackage{hyperref}
\usepackage{amsmath}
\usepackage{amssymb}
\usepackage{siunitx}
\usepackage{tabularx}
\usepackage{array}
\usepackage{geometry}
\makeatletter
\newcommand{\printfnsymbol}[1]{%
  \textsuperscript{\@fnsymbol{#1}}%
}
\makeatother

\begin{document}
\title{Toward Equitable Access: Leveraging Crowdsourced Reviews to Investigate Public Perceptions of Health Resource Accessibility}
\setcounter{footnote}{0}

\author{Zhaoqian Xue\thanks{These authors contributed equally to this work.}\inst{1} \and
Guanhong Liu\printfnsymbol{1}\inst{2} \and
Chong Zhang\inst{3} \and
Kai Wei\inst{4} \and
Qingcheng Zeng\inst{5} \and
Songhua Hu\inst{6} \and
Wenyue Hua\inst{7} \and
Lizhou Fan\inst{8} \and
Yongfeng Zhang\inst{9} \and
Lingyao Li\thanks{Corresponding author.}\inst{10}}

\authorrunning{Z. Xue et al.}

\institute{University of Pennsylvania, Philadelphia, PA, USA\\
\email{Zhaoqian.Xue@pennmedicine.upenn.edu} \and
Renmin University of China, Haidian District, Beijing, China\\
\email{guanhongliu@ruc.edu.cn} \and
University of Liverpool, Liverpool, Merseyside, UK\\
\email{C.Zhang118@liverpool.ac.uk} \and
University of Michigan, Ann Arbor, Michigan, USA\\
\email{weikai@umich.edu} \and
Northwestern University, Evanston, IL, USA\\
\email{qingchengzeng2027@u.northwestern.edu} \and
Massachusetts Institute of Technology, Cambridge, MA, USA\\
\email{hsonghua@mit.edu} \and
University of California, Santa Barbara, Santa Barbara, California, USA\\
\email{wenyuehua@ucsb.edu} \and
Harvard Medical School, Boston, MA, USA\\
\email{lfan8@bwh.harvard.edu} \and
Rutgers University, New Brunswick, New Jersey, USA\\
\email{yongfeng.zhang@rutgers.edu} \and
University of South Florida, Tampa, Florida, USA\\
\email{lingyaol@usf.edu}}
\maketitle              % typeset the header of the contribution
\begin{abstract}
Monitoring health resource disparities during public health crises is critical, yet traditional methods, like surveys, lack the requisite speed and spatial granularity. This study introduces a novel framework that leverages: 1) crowdsourced Google Maps reviews (2018–2021) and 2) advanced NLP (DeBERTa) to create a high-resolution, spatio-temporal index of public perception of health resource accessibility in the United States. We then employ Partial Least Squares (PLS) regression to link this perception index to a range of socioeconomic and demographic drivers. Our results quantify significant spatial-temporal shifts in perceived access, confirming that disparities peaked during the COVID-19 crisis and only partially recovered post-peak. We identify political affiliation, racial composition, and educational attainment as primary determinants of these perceptions. This study validates a scalable method for real-time health equity monitoring and provides actionable evidence for interventions to build a more resilient healthcare infrastructure.
% The accompanying code is available at: \href{https://github.com/Toward-Equitable-Access/Health-Resource}{https://github.com/Toward-Equitable-Access/Health-Resource}.

\end{abstract}
\section{Introduction}

Equitable access to essential health resources is fundamental to public well-being, particularly during health crises when demand surges for medical services and preventive care \cite{who2020}. The distribution of critical supplies—such as medications, personal protective equipment, and testing kits—is vital for controlling disease spread and minimizing mortality \cite{sodhi2023research}. However, significant disparities in health resource availability persist, often exacerbating existing social and economic inequalities \cite{detels2021socioeconomic}.

Traditional methods for assessing health resource accessibility, such as surveys and administrative records, offer valuable insights but are often constrained by significant time lags, high costs, and limited spatial granularity \cite{keppel2005methodological}. These limitations hinder the ability to form timely interventions during rapidly evolving crises. The rise of digital platforms has introduced crowdsourced data as a high-resolution alternative \cite{kim2017problems}. Platforms like Google Maps, in particular, provide granular, real-time, and geo-tagged insights into how the public experiences and perceives access to health resources at the community level \cite{jia2021online}.

This study leverages this novel data source to analyze social inequality by using Google Maps reviews (2018–2021) to investigate how public perceptions of health resource accessibility varied across the U.S. during the COVID-19 pandemic. Specifically, we aim to answer three questions:

\begin{itemize} 
     \item[\textbf{RQ1:}] Do health resource disparities, as perceived by the public, exist across different regions in the United States? \item[\textbf{RQ2:}] Are these perceived health resource disparities correlated with the socioeconomic or demographic characteristics of local communities? \item[\textbf{RQ3:}] Did the COVID-19 pandemic exacerbate health resource disparities as perceived by the public?\end{itemize}

To address these questions, we apply state-of-the-art Natural Language Processing (NLP) techniques, specifically DeBERTa, for text classification and Partial Least Squares (PLS) regression to examine the relationship between perceived accessibility and socioeconomic factors. Our findings reveal significant spatial and temporal disparities in public perception, which peaked during the crisis and eased only partially afterward. We identify political affiliation, racial composition, and educational attainment as key drivers of these perceptions. These results highlight the utility of crowdsourced data for monitoring health equity in real-time and underscore the need for targeted policies to build a more resilient and equitable healthcare infrastructure.

\section{Related Work}

Traditional studies of healthcare access, which rely on surveys and administrative data, often lack the temporal resolution and spatial granularity required to track dynamic public health challenges \cite{gao2016assessment}. In contrast, crowdsourced digital data offers large-scale, passive, and real-time insights into public experiences \cite{wazny2017crowdsourcing}. Advancements in natural language processing (NLP) and machine learning have further enhanced our ability to systematically analyze this unstructured user-generated content \cite{khan2023exploring}.

\setlength{\intextsep}{0pt}
\begin{wraptable}{r}{0.475\textwidth}
\centering
\caption{Keyword ontology for health resources}
\label{tab:keywords}
\begin{tabular}{>{\raggedright\arraybackslash}p{3cm}p{4.5cm}}
\toprule
\textbf{Category} & \textbf{Keywords} \\
\midrule
Essential health supplies & sanitizer, soap, toilet paper, mask, disinfectant, gloves, thermometer, tissues, wipes, face shield, hand wash, respirators, alcohol \\
\midrule
Over-the-counter medications & acetaminophen, tylenol, advil, motrin, ibuprofen, dayquil, nyquil, mucinex, robitussin, sudafed, pepto-bismol, tums, vick's vaporub \\
\midrule
Preventive healthcare items & vitamins, zinc, pedialyte, gatorade \\
\midrule
Diagnostic tools & test kit, home test, self test \\
\midrule
COVID-19 specific items & N95, hydroxychloroquine \\
\midrule
Household sanitization products & lysol spray, disinfectant wipes \\
\bottomrule
\end{tabular}
\end{wraptable}

Previous research has leveraged crowdsourced data, particularly from social media, to monitor disease outbreaks \cite{gui2017managing} and track public sentiment on vaccines \cite{broniatowski2018weaponized}. However, social media data generally lacks the precise, location-specific context needed to assess health resource accessibility at a community level \cite{zohar2021geolocating}. Online review platforms, such as Google Maps, overcome this limitation by providing richer, geo-tagged insights into real-world experiences with specific healthcare providers \cite{ranard2016yelp}. Despite this clear advantage, the potential of these platforms to capture dynamic spatial-temporal disparities and evolving public perceptions of health access remains critically underexplored.

This study addresses this gap by leveraging Google Maps reviews to analyze spatial and temporal disparities in perceived health resource accessibility. Our approach provides a scalable, real-time complement to traditional surveillance and a significant enhancement over less granular crowdsourced data. The findings offer actionable, data-driven insights for policymakers and public health officials, enabling the development of more equitable and responsive health policies.

\section{Dataset and Methodology}

This study's methodology involved four primary stages. First, we acquired and prepared a large-scale corpus of Google Maps reviews for U.S. stores spanning 2018 to 2021 (Section \ref{sec:dataset}). Second, we applied a keyword filtering process to isolate reviews relevant to health resource accessibility (Section \ref{sec:keywords}). Third, we developed and validated text classification models using natural language processing (NLP) to categorize public perceptions (Section \ref{sec:textcla}). Finally, we aggregated these classifications into perception scores (Figure \ref{fig:classfiers}) and employed Partial Least Squares (PLS) regression to analyze their relationship with regional socioeconomic and demographic factors across different time periods (Section \ref{sec:pls}).

\subsection{Data Collection} \label{sec:dataset}

This study utilized the large-scale Google Local Data dataset \cite{yan2023personalized}, which contains 666,324,103 reviews for 4,963,111 U.S. businesses up to September 2021. This dataset, provided in a one-review-per-line JSON format, includes detailed review text, business metadata, and related links.

We selected this data source as it provides a rich, geo-tagged corpus of user-generated content, ideal for analyzing spatio-temporal perceptions of public-facing services \cite{mehta2019google}. For the purpose of this study, we extracted all reviews posted between January 2018 and December 2021 to form our initial analysis corpus.

To capture the pandemic's dynamic impact on public perception, we segmented the dataset into three periods. The Pre-Pandemic Period (January 1, 2018 – January 31, 2020) serves as a baseline. The Peak-Pandemic Period (February 1, 2020 – May 31, 2020) begins with the WHO's declaration of a global health emergency \cite{Cucinotta2020}. This timeframe covers the acute phase of health supply shortages \cite{trump2020proclamation} and federal interventions, such as the Defense Production Act \cite{peters2020dpa,trump2020dpa}, concluding as supply chains began to stabilize by late May 2020 \cite{gao2021report}. Finally, the Post-Peak Period (June 1, 2020 – May 31, 2021) spans the subsequent year until our data collection concluded. This temporal segmentation enables a nuanced analysis of perception shifts across the distinct phases of the crisis.

\vspace{\baselineskip}

\begin{table*}[htbp]
\centering
\caption{Representative Google Maps reviews for public perception of health resource accessibility}
\label{tab:google-reviews}
\small

\setlength{\tabcolsep}{8pt}

\begin{tabularx}{\textwidth}{
    >{\raggedright\arraybackslash\hsize=0.45\hsize}X
    >{\raggedright\arraybackslash\hsize=0.35\hsize}X
    >{\raggedright\arraybackslash\hsize=0.15\hsize}X
}
\toprule
\textbf{Google Maps review} & \textbf{Targeted text} & \textbf{Attitude} \\
\midrule
Just absolutely crazy! There there was no hamburger and no toilet paper, and not hardly no potato chips on the shelves. People were grabbing up stuff like this was the end of the world. & There there was no hamburger and no toilet paper... & Shortage \\
\midrule
Ran out of a lot of paper goods - toilet paper, paper towels. Ended up buying the more expensive paper towels and toilet paper, which I really could not afford, but I needed it. & Ran out of a lot of paper goods - toilet paper, paper towels. & Shortage \\
\midrule
I was able to buy toilet paper there at the height of the shortage, and they sanitized every cart before use. & I was able to buy toilet paper there at the height of the shortage... & No Shortage \\
\midrule
I was thankfully able to visit the Dollar General store location off of Donaghey, and they had plenty of paper products, as well as hand soaps. & ...and they had plenty of paper products, as well as hand soaps. & No Shortage \\
\midrule
The only major store that requires you to wear a mask to shop, but the employees near the entrance having masks hanging below their nose makes it pointless to wear a mask. & The only major store that requires you to wear a mask to shop...having masks hanging below their nose makes it pointless to wear a mask. & Unrelated \\
\midrule
Boards are in place between customers and employees at the payment counter to encourage social distancing, but then only 1 out of the 3 staff present seem to know how to wear a mask properly. & ...but then only 1 out of the 3 staff present seem to know how to wear a mask properly. & Unrelated \\
\bottomrule
\end{tabularx}
\setlength{\tabcolsep}{6pt}
\end{table*}

\vspace{\baselineskip}

To enhance analytical robustness, we integrated three U.S. Census Bureau datasets. First, we used the Household Pulse Survey \cite{census2021pulse}—a survey tracking COVID-19's socioeconomic household impacts—to validate our crowdsourced perception scores against broader population experiences \cite{danek2023users}. Second, administrative division shapefiles \cite{census2023tigerline} provided the precise geographic boundaries necessary for defining our spatial analysis units. Third, socioeconomic data \cite{census2023tiger} served as the independent variables in our regression modeling, enabling us to identify key demographic drivers (e.g., income, education, race) of perceived accessibility \cite{marmot2005social}.

% \begin{table}[htbp]
% \centering
% \caption{Keyword ontology for health resources}
% \label{tab:keywords}
% \centering
% \begin{tabular}{>{\raggedright\arraybackslash}p{3cm}p{4.5cm}}
% \toprule
% \textbf{Category} & \textbf{Keywords} \\
% \midrule
% Essential health supplies & sanitizer, soap, toilet paper, mask, disinfectant, gloves, thermometer, tissues, wipes, face shield, hand wash, respirators, alcohol \\
% \midrule
% Over-the-counter medications & acetaminophen, tylenol, advil, motrin, ibuprofen, dayquil, nyquil, mucinex, robitussin, sudafed, pepto-bismol, tums, vick's vaporub \\
% \midrule
% Preventive healthcare items & vitamins, zinc, pedialyte, gatorade \\
% \midrule
% Diagnostic tools & test kit, home test, self test \\
% \midrule
% COVID-19 specific items & N95, hydroxychloroquine \\
% \midrule
% Household sanitization products & lysol spray, disinfectant wipes \\
% \bottomrule
% \end{tabular}
% \end{table}

\subsection{Keywords ontology and development}
\label{sec:keywords}

To filter reviews relevant to health resources, we developed a comprehensive keyword ontology. The process began with foundational terms identified by the CDC and WHO as critical during health crises (e.G., sanitizer,'' mask,'' thermometer'') \cite{liang2020development}, items known for acute shortages \cite{kass2020obesity}. This core list was expanded to capture consumer-specific language, including brand-name medications (Tylenol,'' Advil''), household products (Lysol spray'') \cite{keller2020consumer}, and emergent pandemic terminology (N95,'' home test'') \cite{andrejko2022effectiveness}. We also incorporated highly publicized, controversial terms (e.g., ``hydroxychloroquine'') to reflect media-driven public discourse \cite{ben2012hydroxychloroquine,rea2021open,borriello1890liver,self2020effect}. A complete list is provided in Table~\ref{tab:keywords}.

This development process was iterative and validated against existing consumer health studies and social media analyses to ensure contextual relevance and comprehensiveness \cite{awan2023neoteric,amur2023unlocking}. Applying this finalized ontology yielded our analytical dataset of 289,919 reviews.

\setlength{\intextsep}{0pt}
\begin{wrapfigure}{r}{0.475\textwidth}
    \centering
    \includegraphics[width=0.475\textwidth]{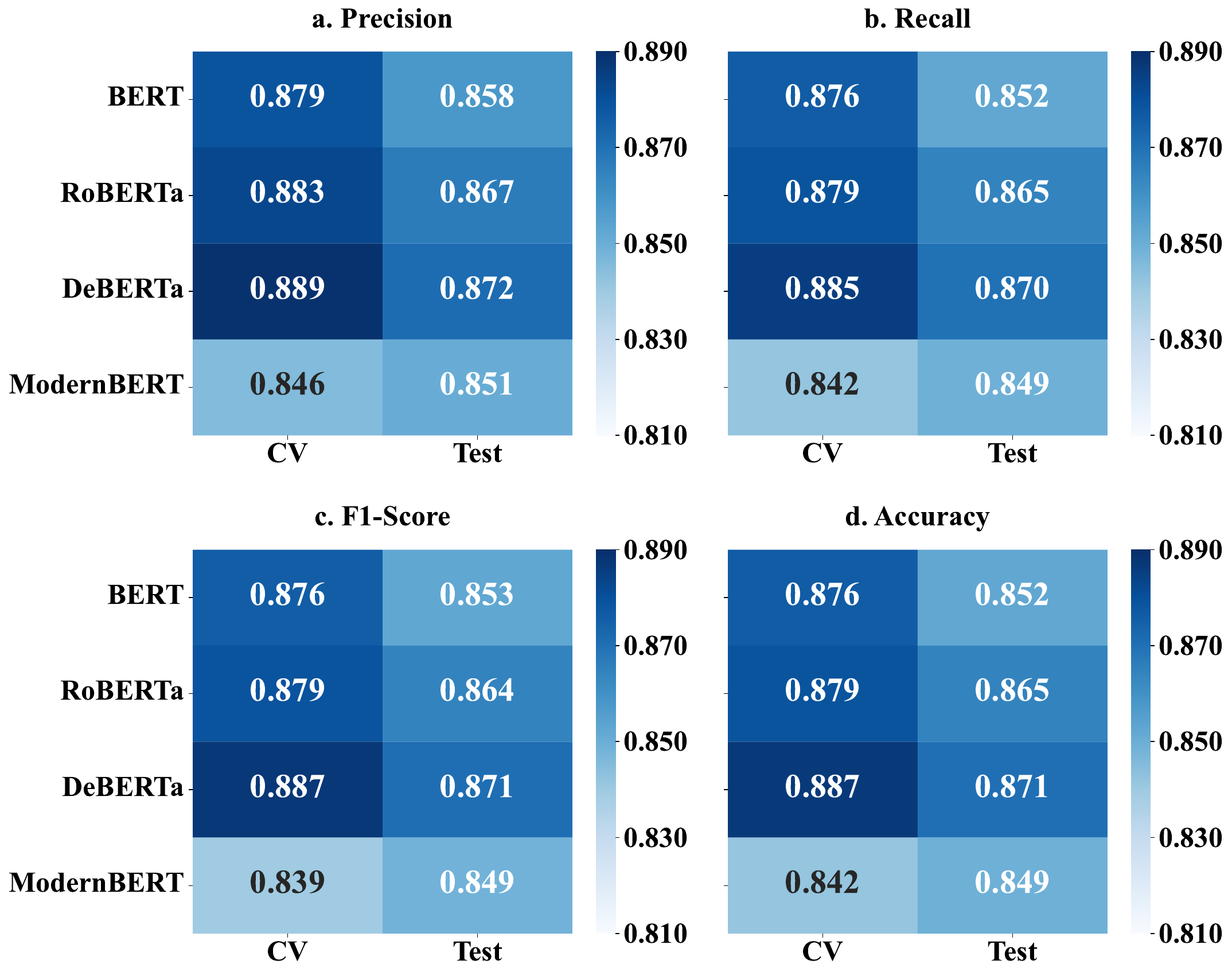}
    \vspace{-10pt}
    \caption{Classification performance of candidate models: (a) Precision, (b) Recall, (c) F1-score, and (d) Training and testing accuracy.}
    \label{fig:classfiers}
    \vspace{10pt}
\end{wrapfigure}

\subsection{Text classification}
\label{sec:textcla}

We developed a text classification model to categorize reviews into three classes: Class -1 (shortage of health resources), Class 1 (no shortage), and Class 9 (unrelated). To create a balanced annotation dataset, we used a sequential labeling process until each of the three classes contained approximately 500 instances (1,500 reviews total). This dataset was annotated by three public health experts; each review was labeled independently by two experts, achieving a Cohen’s Kappa of 0.85 \cite{cohen1960coefficient}. A third expert resolved all disagreements. The final dataset was split 80\% for training and 20\% for testing.

We vectorized the text using transformer-based embeddings and evaluated four models: BERT \cite{devlin2019bert}, RoBERTa \cite{liu2019roberta}, DeBERTa \cite{he2020deberta}, and ModernBERT \cite{warner2024smarter}. After hyperparameter tuning using 5-fold cross-validation and grid search \cite{kohavi1995study}, models were evaluated on Accuracy, Precision, Recall, and F1-Score \cite{powers2020evaluation}. As shown in Figure~\ref{fig:classfiers}, DeBERTa consistently outperformed other models, achieving a test accuracy of 87.05\% and an F1-Score of 0.8709. We therefore selected DeBERTa as our final classifier. We applied this trained model to classify all 289,919 reviews, retaining only those identified as Class -1 or Class 1 for the subsequent analysis.

To quantitatively evaluate public perceptions, we calculated a public perception score from the model's classification outcomes. This score served as a proxy for the collective community experience regarding health resource access across different regions and time periods. We defined the perception score $S_{c,t}$ for each county $c$ and time period $t$ as the average of the numerical labels assigned to the classified reviews:
\begin{equation}
    S_{c,t} = \frac{1}{N_{c,t}} \sum_{i=1}^{N_{c,t}} L_i
\end{equation}
where $N_{c,t}$ is the total number of reviews in county $c$ during time period $t$ that were classified as either indicating a shortage or no shortage (Class -1 or Class 1), and $L_i$ is the numerical label assigned to each review $i$, with $L_i = -1$ for Class -1 (indicating a shortage) and $L_i = 1$ for Class 1 (indicating no shortage).

The score is the arithmetic mean of the classification labels (where $-1$ = Shortage, $1$ = No Shortage) for all relevant reviews within that spatio-temporal unit. This score ranges from $-1$ (indicating a prevalence of perceived shortages) to $1$ (indicating perceived abundance), with $0$ representing a balanced perception.

To ensure the robustness of this metric, our analysis only included counties with a minimum of 10 relevant reviews ($L_i = -1$ or $1$) within a given time period (Section~\ref{sec:dataset}). This minimum threshold aligns with standard practices in social media analytics to ensure data sufficiency and enhance the generalizability of findings \cite{faber2014sample,mason2010sample}.

\subsection{Partial Least Squares (PLS) Regression}
\label{sec:pls}

\setlength{\intextsep}{0pt}
\begin{wraptable}{r}{0.45\textwidth}
\centering
\caption{VIF for features in the model}
\label{tab:vif_table}
\begin{tabular}{lc}
\toprule
\textbf{Feature} & \textbf{VIF} \\
\midrule
Democratic Rate & 39.272 \\
Republican Rate & 36.343 \\
Total Population & 1.637 \\
Median Income & 10.778 \\
GINI & 3.759 \\
No Insurance Rate & 2.848 \\
Household Below Poverty Rate & 9.395 \\
HISPANIC LATINO Rate & 5.455 \\
White Rate & 22.867 \\
Black Rate & 18.637 \\
Indian Rate & 1.822 \\
Asian Rate & 7.005 \\
Under 18 Rate & \textit{inf} \\
Between 18 and 44 Rate & \textit{inf} \\
Between 45 and 64 Rate & \textit{inf} \\
Over 65 Rate & \textit{inf} \\
Male Rate & 2.140 \\
Bachelor Rate & 16.745 \\
Education Degree Rate & 28.655 \\
Population Density & 1.313 \\
Unemployed Rate & 3.353 \\
\bottomrule
\end{tabular}
\end{wraptable}

We employed Partial Least Squares (PLS) regression at the county level ($n=530$) to analyze the relationship between perception scores and socioeconomic factors. This method was specifically chosen to address significant multicollinearity among independent variables—such as the Democratic and Republican rates—which exhibited Variance Inflation Factors (VIF) exceeding 5 ~\ref{tab:vif_table}. Unlike variable removal, which risks omitting key predictors, PLS regression effectively manages collinearity by decomposing both predictor and response variables into orthogonal scores and loadings \cite{deJong1993simpls}. The key equations of PLS regression are presented below:

\begin{equation}
    \begin{array}{l}
        X=T P^T+E \\
        Y=U Q^T+F \\
        Y=X K^T+\Theta
    \end{array}
\end{equation}

\noindent where independent variables $X$ represent the socioeconomic data hypothesized to correlate with public perceptions. The dependent variable $Y$ was modeled in five ways: the average perception scores for each of the three individual periods (RQ2), and the change in scores between periods (e.g., Peak-Pandemic vs. Pre-Pandemic) to investigate pandemic-driven shifts (RQ3). All predictor $X$ and response $Y$ variables were standardized as Z-scores. The model was solved using the SIMPLS algorithm \cite{de1993simpls}, and its overall performance was assessed on the full dataset using $R^2$ and Root Mean Square Error (RMSE). As PLS does not inherently provide p-values for coefficients, we employed permutation testing ($n_{\text{perm}} = 1000$) to assess statistical significance. This method generates robust p-values and standard errors by comparing observed coefficients against a null distribution from randomly permuted data \cite{Afthanorhan2015}. The key equations for the permutation tests are:

\begin{equation}
\begin{aligned}
\Theta^{\text{obs}} & = \text{Coefficient derived from the original dataset}, \\
\Theta^{(k)} & = \text{Coefficient derived from the $k$-th permutation}, \\
p_i & = \frac{1}{n_{\text{perm}}} \sum_{k=1}^{n_{\text{perm}}} \mathbb{I}(|\Theta_i^{(k)}| \geq |\Theta_i^{\text{obs}}|), \\
SE(\Theta_i) & = \sqrt{\frac{\sum_{k=1}^{n_{\text{perm}}} (\Theta_i^{(k)} - \bar{\Theta}_i)^2}{n_{\text{perm}} - 1}}.
\end{aligned}
\end{equation}

\noindent where $\Theta^{\text{obs}}$ represents the observed regression coefficients from the original dataset. $\Theta^{(k)}$ denotes the coefficients derived from the $k$-th permutation of the dependent variable $Y$, with $n_{\text{perm}}$ set to 1000 in this study. The $p$-value ($p_i$) for each coefficient is the proportion of permuted coefficients ($|\Theta_i^{(k)}|$) that are as or more extreme than the observed coefficient ($|\Theta_i^{\text{obs}}|$). The standard error $SE(\Theta_i)$ is calculated as the standard deviation of the permuted coefficients $\Theta_i^{(k)}$ around their mean $\bar{\Theta}_i$.

In addition, the standard error $SE(\Theta_i)$ is calculated as the standard deviation of the permuted coefficients $\Theta_i^{(k)}$ around their mean $\bar{\Theta}_i$.

\section{Results}

This section presents our results, structured around our three research questions. First (RQ1), we analyze the spatial distribution of public perception scores, visualizing geographic patterns and quantifying clustering using Moran's I (Section \ref{lab:RQ1}). Second (RQ2), we use PLS regression to investigate the relationship between county-level socioeconomic characteristics and perception scores across the Pre-Pandemic, Peak-Pandemic, and Post-Peak periods (Section \ref{lab:RQ2}). Third (RQ3), we evaluate whether the pandemic exacerbated disparities by analyzing the change in perception scores between these periods (Section \ref{lab:RQ3}).

\begin{wrapfigure}{r}{0.45\textwidth}
    \centering
    \includegraphics[width=0.45\textwidth]{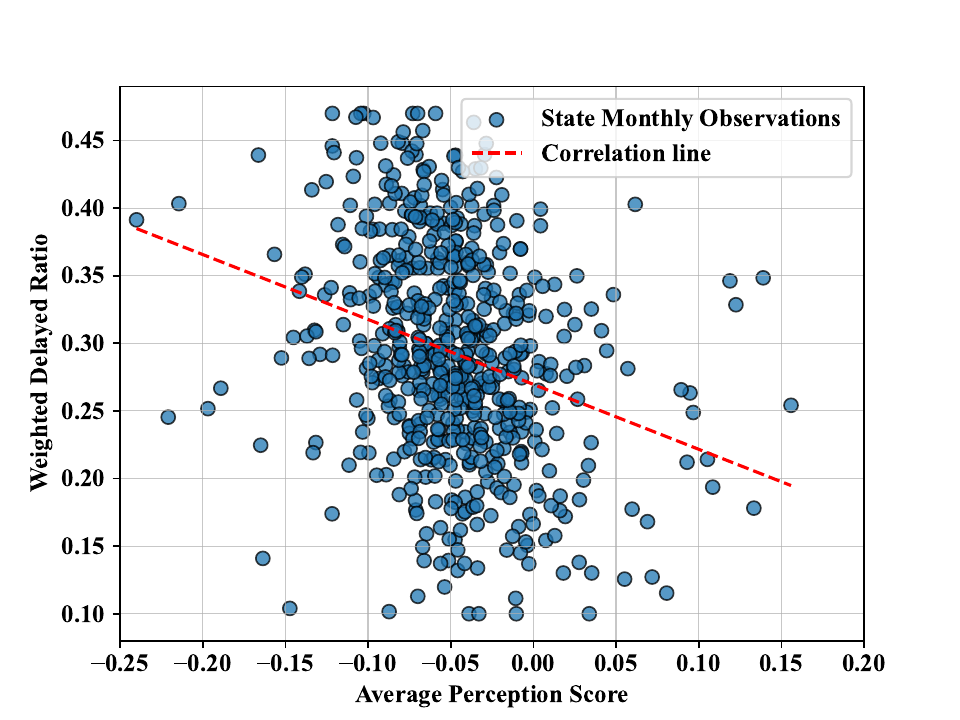}
    \caption{Comparison between weighted delayed ratio reported by US Census Bureau and average perception scores by online reviews.}
    \label{fig:validation}
\end{wrapfigure}

Before this analysis, we first validated our metric. We compared our monthly, state-level perception scores against the U.S. Census Bureau's Household Pulse Survey \cite{census2021pulse} from April 2020 to April 2021 (Figure \ref{fig:validation}). We calculated the correlation between our aggregated scores and the proportion of survey respondents reporting delays in securing health resources, confirming that our crowdsourced measure aligns with established national survey data \cite{danek2023users}.

\setlength{\intextsep}{0pt}

To ensure robust parameter estimates and identify observations that were both outliers and high-leverage points, which could significantly distort the true correlation trend \cite{psu462}, we employed Cook’s Distance \cite{Nurunnabi2015}. For an observation $i$, Cook's Distance is calculated as:

\begin{equation}
D_i = \frac{\sum_{j=1}^n \left( \hat{y}_j^{(-i)} - \hat{y}_j \right)^2}{p \cdot \text{MSE}}
\end{equation}

\noindent where: $ \hat{y}_j^{(-i)} $ is the predicted value for observation $ j $ when observation $ i $ is excluded, $ \hat{y}_j $ is the predicted value using the full dataset, $ p $ is the number of model parameters (including the intercept), and $ \text{MSE} $ is the mean squared error of the regression model. To identify influential observations, we applied a threshold defined as:
\begin{equation}
\text{Threshold} = \frac{4}{n}
\end{equation}
\noindent

\noindent where $ n $ is the total number of observations in the dataset. Observations with $ D_i > \text{Threshold} $ were flagged as extreme observations, and they will be removed.

\setlength{\intextsep}{0pt}
\begin{wrapfigure}{r}{0.45\textwidth}
    \centering
    \includegraphics[width=0.45\textwidth]{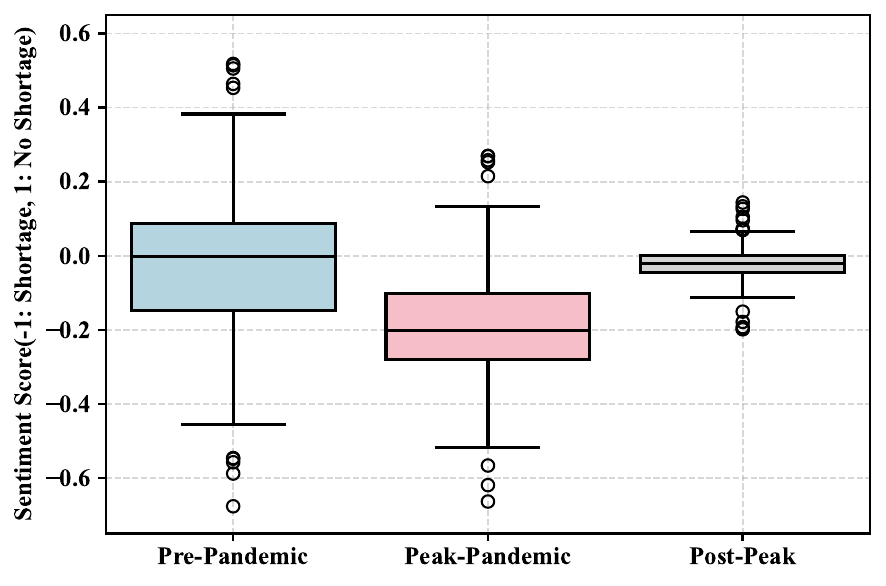}
    \caption{Health resource availability trends across pandemic periods.}
    \label{fig:sentiment}
\end{wrapfigure}

As shown in Figure \ref{fig:validation}, the red dashed line shows the trend with a negative correlation ($r$ = -0.2786, $p$ = \num{3.73e-12}). This suggests that states with higher public perception scores of health resource accessibility tended to have fewer reported delays in securing health resources. Although the correlation is slight, it implies that online reviews reflect the situation in terms of health resource accessibility.

\subsection{RQ1: Do health resource disparities, as perceived by the public, exist across different regions in the United States?}
\label{lab:RQ1}

\begin{wrapfigure}{r}{0.5\textwidth}
    \centering
    \includegraphics[width=0.5\textwidth]{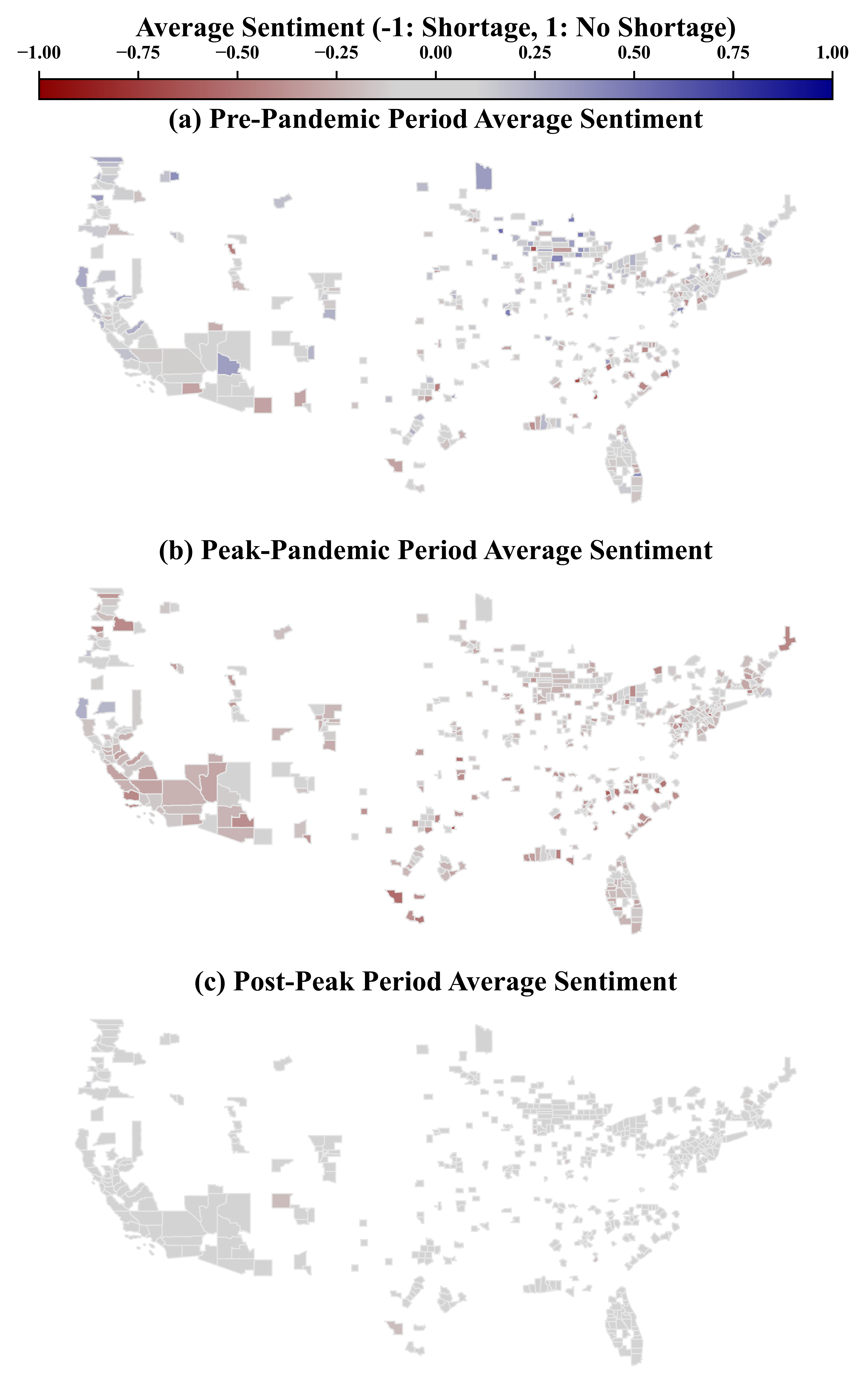}
    \caption{Geographic patterns of the perceived health resource disparities across the United States.}
    \label{fig:sentiment_maps}
\end{wrapfigure}

To address RQ1, we analyzed county-level perception scores (ranging from -1, negative, to 1, positive) across the three pandemic periods. As illustrated in Figure~\ref{fig:sentiment}, the average perception score dropped significantly during the Peak-Pandemic period and remained low throughout the Post-Peak period. The geographic distribution of these scores is visualized in Figure~\ref{fig:sentiment_maps}, where red denotes negative perceptions (shortages) and blue represents positive perceptions (availability).

During the Pre-Pandemic period (Figure~\ref{fig:sentiment_maps}a), negative perceptions were particularly pronounced across parts of the Western states, while the Eastern regions exhibited a more nuanced mix of positive and negative perceptions, especially in densely populated areas. During the Peak-Pandemic period (Figure~\ref{fig:sentiment_maps}b), these regional disparities in public perceptions became more pronounced. Western states, notably California and its surrounding areas, experienced a significant surge in negative perceptions, as reflected by larger red clusters. Meanwhile, the Eastern region maintained its heterogeneous pattern but exhibited a noticeable shift toward more negative perceptions. By the Post-Peak period (Figure~\ref{fig:sentiment_maps}c), a substantial transition toward neutral perceptions was observed nationwide, as indicated by the predominance of light gray tones on the map.

\begin{wrapfigure}{r}{0.5\textwidth}
    \centering
    \includegraphics[width=0.5\textwidth]{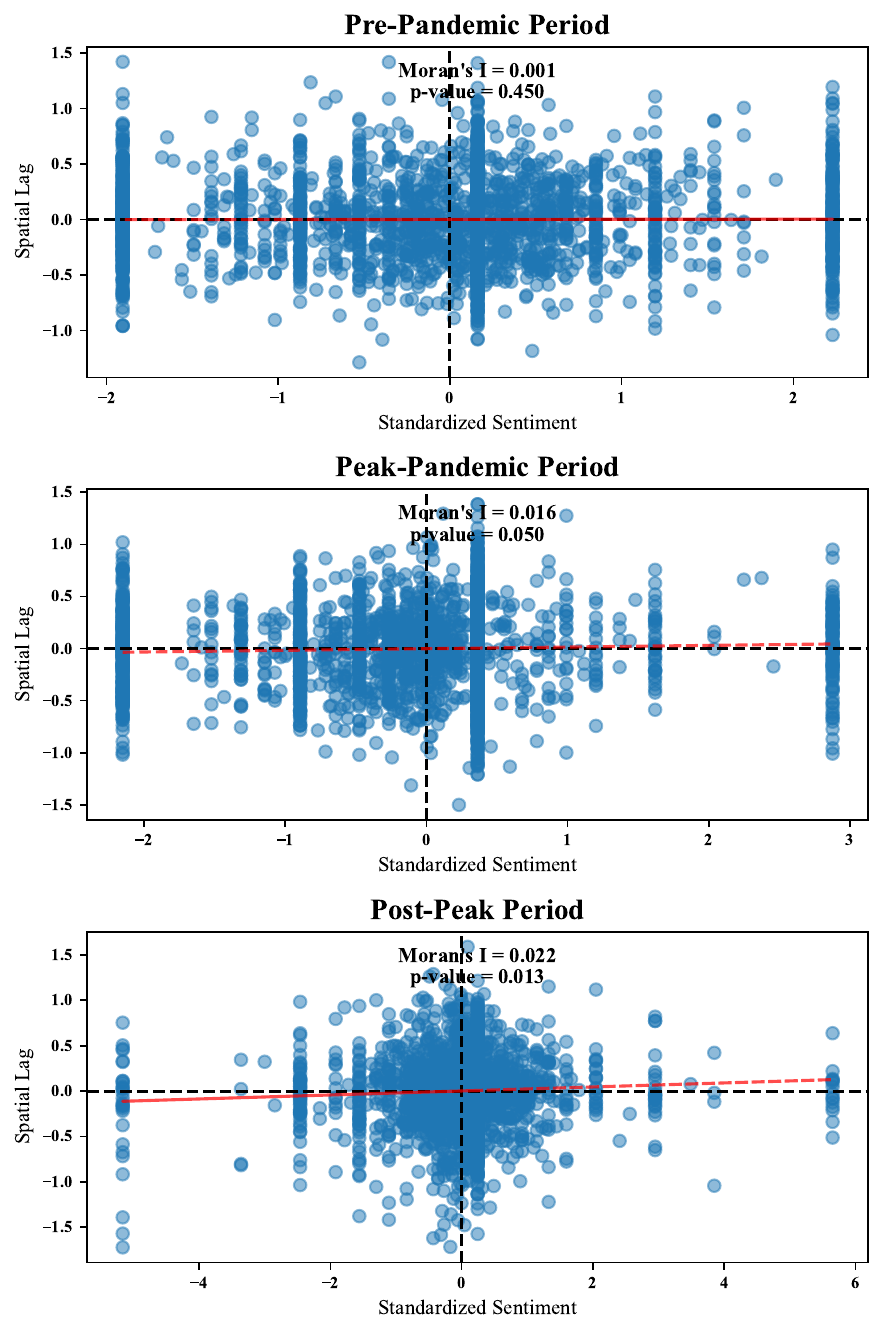}
    \caption{Moran's I scatterplots across pandemic periods.}
    \label{fig:morans_i}
\end{wrapfigure}

We used Moran's I statistic to perform the spatial autocorrelation analysis and quantify the patterns of public perceptions, as shown in Figure~\ref{fig:morans_i}. Before the pandemic, perception scores showed virtually no spatial clustering among neighboring counties, as indicated by a low, non-significant Moran's I ($0.001$, $p = 0.450$). This suggests a relatively random geographical distribution of public perceptions regarding health resource accessibility. As the pandemic reached its peak, however, we observed the emergence of weak yet statistically significant spatial patterns in public perception (Moran's I $= 0.016$, $p = 0.050$). In the Post-Peak period, spatial autocorrelation became more pronounced and statistically significant (Moran's I $= 0.022$, $p = 0.013$), indicating that counties with similar perception scores (either positive or negative) were increasingly clustered geographically. This trend towards heightened spatial clustering, visually corroborated by Figure~\ref{fig:sentiment_maps} where regional concentrations of sentiment become more distinct over time, suggests that the pandemic's impact on health resource accessibility was not geographically uniform. Instead, the formation of these significant clusters---areas where perceptions of shortages were concentrated versus areas where perceptions of adequacy were more common---points to an unmasking or deepening of regional disparities. Such a geographical sorting of experiences implies that the pandemic may have exposed and potentially exacerbated underlying regional inequalities in health resource availability, by making the differences in access between various areas more geographically distinct and patterned.

\setlength{\intextsep}{0pt}

\subsection{RQ2: Are these perceived health resource disparities correlated with the socioeconomic and demographic factors?}
\label{lab:RQ2}

The PLS regression overall revealed a strong correlation between public perceptions and socioeconomic characteristics (Table \ref{tab:regression-results}), with significant influences observed from factors such as race, income inequality, education, and even age. 

Before the outbreak of COVID-19, the White Rate (coeff = 0.026, $p$ < 0.001) exhibited a positive association with perception scores, indicating better access to health resources in counties with a higher proportion of White residents. Conversely, counties with a larger proportion of Black residents (Black Rate: coeff = -0.047, $p$ < 0.001) tended to have less favorable access. Additionally, a higher Republican population rate (Republican Rate: coeff = -0.027, $p$ = 0.023) and a higher uninsured population rate (No Insurance Rate: coeff = -0.022, $p$ < 0.001) were negatively associated with health resource accessibility, highlighting disparities in Republican-leaning regions and among uninsured populations.

\vspace{\baselineskip}
\begin{table*}[htbp]
\centering
\caption{PLS regression results of socioeconomic factors on perception scores across pandemic periods.}
\label{tab:regression-results}
\small
\begin{tabularx}{\textwidth}{>{\raggedright\arraybackslash}Xccccccccc}
\toprule
\textbf{Variable} & \multicolumn{3}{c}{\textbf{Pre-Pandemic}} & \multicolumn{3}{c}{\textbf{Peak-Pandemic}} & \multicolumn{3}{c}{\textbf{Post-Peak}} \\
\cmidrule(lr){2-4} \cmidrule(lr){5-7} \cmidrule(lr){8-10}
& {Coeffs} & {P-Value} & {Std. err.} & {Coeffs} & {P-Value} & {Std. err.} & {Coeffs} & {P-Value} & {Std. err.} \\
\midrule
Democratic Rate & 0.021 & 0.061 & 0.043 & 0.009 & 0.004** & 0.044 & -0.004 & 0.020* & 0.043 \\
Republican Rate & -0.027 & 0.023* & 0.046 & -0.018 & 0.001** & 0.045 & -0.002 & 0.001** & 0.046 \\
Total Population & 0.003 & 0.720 & 0.045 & 0.007 & 0.790 & 0.046 & -0.002 & 0.610 & 0.044 \\
Median Income & -0.017 & 0.610 & 0.043 & -0.014 & 0.009** & 0.043 & 0.003 & 0.004** & 0.045 \\
GINI & 0.014 & 0.071 & 0.045 & -0.014 & 0.850 & 0.045 & 0.003 & 0.105 & 0.044 \\
No Insurance Rate & -0.022 & <0.001** & 0.045 & -0.024 & <0.001** & 0.046 & -0.006 & <0.001** & 0.045 \\
Household Below Poverty Rate & -0.006 & 0.985 & 0.044 & 0.002 & 0.005** & 0.043 & -0.002 & 0.108 & 0.046 \\
HISPANIC LATINO Rate & -0.022 & 0.060 & 0.046 & -0.020 & <0.001** & 0.046 & 0.003 & 0.270 & 0.043 \\
White Rate & 0.026 & <0.001** & 0.043 & 0.002 & 0.655 & 0.045 & -0.001 & 0.650 & 0.045 \\
Black Rate & -0.047 & <0.001** & 0.046 & -0.007 & 0.255 & 0.044 & -0.002 & 0.370 & 0.043 \\
Indian Rate & 0.004 & 0.510 & 0.044 & 0.016 & 0.135 & 0.046 & -0.007 & 0.008** & 0.045 \\
Asian Rate & 0.006 & 0.690 & 0.043 & -0.002 & 0.060 & 0.045 & 0.002 & 0.001** & 0.043 \\
Under 18 Rate & 0.003 & <0.001** & 0.045 & 0.004 & <0.001** & 0.046 & -0.005 & 0.012* & 0.046 \\
Between 18 and 44 Rate & 0.002 & 0.830 & 0.043 & -0.008 & 0.820 & 0.043 & 0.004 & 0.011* & 0.045 \\
Over 65 Rate & 0.006 & 0.014* & 0.046 & 0.012 & 0.078 & 0.045 & 0.002 & 0.530 & 0.044 \\
Male Rate & -0.018 & 0.885 & 0.044 & -0.011 & 0.105 & 0.044 & -0.001 & 0.950 & 0.043 \\
Bachelor Rate & 0.005 & 0.170 & 0.043 & 0.014 & <0.001** & 0.045 & 0.003 & <0.001** & 0.046 \\
Education Degree Rate & -0.010 & 0.155 & 0.046 & 0.013 & <0.001** & 0.043 & 0.002 & <0.001** & 0.044 \\
Population Density & -0.011 & 0.805 & 0.045 & 0.002 & 0.300 & 0.046 & 0.002 & 0.075 & 0.046 \\
Unemployed Rate & 0.004 & 0.570 & 0.044 & -0.004 & 0.014* & 0.045 & 0.003 & 0.110 & 0.045 \\
\midrule
\textbf{Goodness-of-fit} & \multicolumn{3}{c}{$R^2 = 0.138$, RMSE = 0.169} & \multicolumn{3}{c}{$R^2 = 0.146$, RMSE = 0.123} & \multicolumn{3}{c}{$R^2 = 0.090$, RMSE = 0.037} \\
\bottomrule
\end{tabularx}
\end{table*}

The PLS regression revealed a strong correlation between public perceptions and socioeconomic characteristics (Table \ref{tab:regression-results}), with significant influences observed from factors such as race, income inequality, education, and age.In the Pre-Pandemic period, results confirm significant baseline disparities. Perceptions were more positive in counties with a higher White Rate (coeff = 0.026, $p$ < 0.001) but strongly negative in those with a higher Black Rate (coeff = -0.047, $p$ < 0.001). Negative associations were also found for the Republican Rate (coeff = -0.027, $p$ = 0.023) and the No Insurance Rate (coeff = -0.022, $p$ < 0.001), highlighting pre-existing vulnerabilities.The Peak-Pandemic period revealed a new set of factors. Politically, the Democratic Rate (coeff = 0.009, $p$ = 0.004) was associated with better accessibility, while the Republican Rate (coeff = -0.018, $p$ = 0.001) was negative. Economic factors showed a complex trend: Median Income (coeff = -0.014, $p$ = 0.009) and the Unemployed Rate (coeff = -0.004, $p$ = 0.014) were negatively associated with perceptions, whereas the Household Below Poverty Rate (coeff = 0.002, $p$ = 0.005) showed a slight positive association. Racial disparities shifted, with counties with higher Hispanic/Latino populations (coeff = -0.020, $p$ < 0.001) experiencing significant negative pressure. Finally, educational attainment emerged as a strong positive predictor (Bachelor Rate: coeff = 0.014, $p$ < 0.001; Education Degree Rate: coeff = 0.013, $p$ < 0.001).In the Post-Peak period, the dynamics shifted again. Educational attainment remained a positive, though weaker, factor (Bachelor Rate: coeff = 0.003, $p$ < 0.001). Politically, perceptions became negative in areas associated with both parties, particularly the Democratic Rate (coeff = -0.004, $p$ = 0.020). Economic factors aligned with a recovery narrative: Median Income (coeff = 0.003, $p$ = 0.004) became a positive predictor, while the No Insurance Rate (coeff = -0.006, $p$ < 0.001) remained strongly negative. The recovery was demographically uneven: counties with higher Asian (coeff = 0.002, $p$ = 0.001) and Between 18 and 44 (coeff = 0.004, $p$ = 0.011) populations recovered faster, while those with higher Indian (coeff = -0.007, $p$ = 0.008) and Under 18 (coeff = -0.005, $p$ = 0.012) populations showed slower recovery.\\

\subsection{RQ3: Did the COVID-19 pandemic exacerbate health resource accessibility as perceived by the public?}
\label{lab:RQ3}

The results presented in Tables \ref{tab:pre-peak} and \ref{tab:peak-post} reveal significant shifts in perceived health resource accessibility during the pandemic's peak and subsequent recovery. The analysis of the Pre-to-Peak transition (Table \ref{tab:pre-peak}) indicates that the pandemic exacerbated pre-existing inequalities. Predominantly White communities experienced a pronounced decline in perception scores (coeff = -0.024, $p$ = 0.005), potentially reflecting a higher baseline of access that created a larger margin for perceived deterioration. Conversely, Black communities exhibited greater resilience (coeff = 0.040, $p$ = 0.006), possibly due to an already limited resource baseline with less room for further decline.

Analysis of the Peak-to-Post recovery period Table \ref{tab:peak-post} reveals persistent and uneven recovery trajectories. Demographically, counties with higher Hispanic/Latino Rates showed significant improvement (coeff = 0.023, $p$ < 0.001), while those with higher Indian Rates saw a continued decline or slower recovery (coeff = -0.023, $p$ = 0.020). The No Insurance Rate was positively associated with recovery (coeff = 0.018, $p$ < 0.001), suggesting communities with higher uninsured rates perceived greater improvement during this phase. Political demographics also influenced recovery: counties with higher Republican Rates reported greater improvements (coeff = 0.016, $p$ = 0.018), while those with higher Democratic Rates reported less favorable outcomes (coeff = -0.013, $p$ = 0.042). Finally, educational attainment showed a negative association with recovery; counties with a higher Bachelor Rate (coeff = -0.011, $p$ < 0.001) and Education Degree Rate (coeff = -0.011, $p$ < 0.001) perceived less improvement.These findings demonstrate that the pandemic not only intensified existing disparities but also resulted in uneven recovery trajectories, shaped by a complex interplay of demographic, socioeconomic, and political factors.

\vspace{\baselineskip}
\begin{table*}[htbp]
\centering
\begin{minipage}[t]{0.49\textwidth}
    \centering
    \caption{Changes in PLS regression coefficients between Pre-Pandemic and Peak-Pandemic periods (Peak-Pandemic minus Pre-Pandemic).}
    \label{tab:pre-peak}
    \small
    \begin{tabularx}{\linewidth}{>{\raggedright\arraybackslash}Xccc}
    \toprule
    \textbf{Variable} & \textbf{Coeffs} & \textbf{P-Value} & \textbf{Std. err.} \\ \midrule
    Democratic Rate & -0.012 & 0.950 & 0.044 \\
    Republican Rate & 0.009 & 0.920 & 0.045 \\
    Total Population & 0.004 & 0.650 & 0.046 \\
    Median Income & 0.003 & 0.060 & 0.044 \\
    GINI & -0.028 & 0.080 & 0.045 \\
    No Insurance Rate & -0.002 & 0.780 & 0.046 \\
    Household Below Poverty Rate & 0.008 & 0.100 & 0.044 \\
    HISPANIC LATINO Rate & 0.002 & 0.290 & 0.045 \\
    White Rate & -0.024 & 0.005** & 0.044 \\
    Black Rate & 0.040 & 0.006** & 0.045 \\
    Indian Rate & 0.012 & 0.590 & 0.046 \\
    Asian Rate & -0.008 & 0.390 & 0.045 \\
    Under 18 Rate & 0.001 & 0.230 & 0.043 \\
    Between 18 and 44 Rate & -0.010 & 0.760 & 0.044 \\
    Between 45 and 64 Rate & 0.002 & 0.380 & 0.046 \\
    Over 65 Rate & 0.006 & 0.330 & 0.043 \\
    Male Rate & 0.007 & 0.260 & 0.045 \\
    Bachelor Rate & 0.009 & 0.078 & 0.044 \\
    Education Degree Rate & 0.023 & 0.065 & 0.043 \\
    Population Density & 0.013 & 0.410 & 0.045 \\
    Unemployed Rate & -0.008 & 0.330 & 0.044 \\ 
    \midrule
    \textbf{Model Goodness-of-fit} & \multicolumn{3}{c}{$R^2 = 0.071$, RMSE = 0.205} \\
    \bottomrule
    \end{tabularx}
\end{minipage}
\hfill 
\begin{minipage}[t]{0.49\textwidth}
    \centering
    \caption{Changes in PLS regression coefficients between Peak-Pandemic and Post-Peak periods (Post-Peak minus Peak-Pandemic).}
    \label{tab:peak-post}
    \small
    \begin{tabularx}{\linewidth}{>{\raggedright\arraybackslash}Xccc}
    \toprule
    \textbf{Variable} & \textbf{Coeffs} & \textbf{P-Value} & \textbf{Std. err.} \\ \midrule
    Democratic Rate & -0.013 & 0.042* & 0.044 \\
    Republican Rate & 0.016 & 0.018* & 0.043 \\
    Total Population & -0.009 & 0.910 & 0.045 \\
    Median Income & 0.017 & 0.085 & 0.044 \\
    GINI & 0.017 & 0.520 & 0.045 \\
    No Insurance Rate & 0.018 & <0.001** & 0.046 \\
    Household Below Poverty Rate & -0.004 & 0.028* & 0.043 \\
    HISPANIC LATINO Rate & 0.023 & <0.001** & 0.045 \\
    White Rate & -0.003 & 0.590 & 0.046 \\
    Black Rate & 0.005 & 0.360 & 0.045 \\
    Indian Rate & -0.023 & 0.020* & 0.045 \\
    Asian Rate & 0.004 & 0.270 & 0.045 \\
    Under 18 Rate & -0.009 & 0.002** & 0.046 \\
    Between 18 and 44 Rate & 0.012 & 0.350 & 0.046 \\
    Between 45 and 64 Rate & 0.010 & 0.055 & 0.044 \\
    Over 65 Rate & -0.010 & 0.068 & 0.045 \\
    Male Rate & 0.010 & 0.135 & 0.046 \\
    Bachelor Rate & -0.011 & <0.001** & 0.044 \\
    Education Degree Rate & -0.011 & <0.001** & 0.044 \\
    Population Density & 0.000 & 0.990 & 0.046 \\
    Unemployed Rate & 0.007 & 0.065 & 0.045 \\ 
    \midrule
    \textbf{Model Goodness-of-fit} & \multicolumn{3}{c}{$R^2 = 0.122$, RMSE = 0.125} \\
    \bottomrule
    \end{tabularx}
\end{minipage}
\end{table*}
\vspace{\baselineskip}

\section{Discussion}

This study, leveraging crowdsourced Google Maps reviews, reveals significant spatial, socioeconomic, and temporal disparities in public perceptions of health resource accessibility across the United States. Our findings offer granular insights into health equity, particularly highlighting how a public health crisis interacts with pre-existing structural factors.

\subsection{Principal Findings and Implications} Our analysis confirms that perceived health resource accessibility is not uniform; it is shaped by geography, socioeconomic status, and demographic factors. The county-level variations, which shifted notably during the COVID-19 pandemic, underscore the need for regional strategies rather than one-size-fits-all policies.

Furthermore, our findings quantitatively affirm the pandemic's role as an inequality amplifier, consistent with existing crisis literature \cite{connor2020health}. The pandemic exacerbated challenges for communities with higher poverty rates and significant Hispanic/Latino or Asian populations. The uneven post-pandemic recovery—with Hispanic/Latino populations showing improvements while Indian communities recovered slower—demands tailored interventions.

Interestingly, our temporal analysis revealed different impacts across racial groups. The sharper decline in perception scores among White communities may suggest that previously better-resourced areas were more strained by the crisis. Conversely, the relative stability in perceptions among Black communities might reflect a chronic, pre-existing baseline of limited access, leaving less "room" for perceived deterioration from an already low baseline. Our findings linking persistent disparities to insurance status and education level reinforce calls for policy addressing these social determinants of health \cite{hahn2015education,khairat2019advancing}.

\setlength{\intextsep}{0pt}

\subsection{Limitations and Data Representativeness} Acknowledging these implications requires a careful understanding of our data's limitations. Google Maps reviews are a self-selected sample, not a systematically sampled cohort, and may therefore introduce biases.

Two primary forms of bias are relevant. First, demographic bias is a critical concern for health equity. Reviewers may overrepresent urban, younger, and tech-savvy individuals, thereby underrepresenting vulnerable groups (e.g., elderly, rural, low-income) whose perceptions might be inadequately captured, as suggested by sparse data in some regions (Figure~\ref{fig:sentiment_maps}). Second, reporting bias, such as a negativity bias where dissatisfied users are more vocal, could overrepresent negative sentiment.

Although our focus on relative differences (rather than absolute scores) may mitigate some consistent underlying bias, and our scores showed modest correlation with the Household Pulse Survey (Figure \ref{fig:validation}), these findings must be interpreted with caution.

\subsection{Future Research Directions} Future work should directly address these limitations. The most critical step is to mitigate representation bias through methodological enhancements and data fusion. This includes integrating crowdsourced data with traditional surveys or administrative records to create composite indices that are both granular and representative. Developing advanced bias correction techniques or weighting schemes for these novel data streams would be a valuable contribution.

Additionally, classification performance could be enhanced by moving beyond traditional models. Using more advanced large language models (LLMs), trained on expanded and robustly annotated datasets, could yield more fine-grained and accurate insights into public perceptions of healthcare.

\section{Conclusions}
This study demonstrates the value of crowdsourced data, such as Google Maps reviews, in identifying disparities in public perceptions of health resource accessibility. During the COVID-19 pandemic, these disparities intensified at the peak and showed slight improvement in the Post-Peak period, influenced by socioeconomic and demographic factors. Geospatial and regression analyses indicate more positive public perceptions among White, insured, wealthy, and educated communities. The findings underscore the need for targeted policies to address inequities and build more inclusive, resilient communities to withstand future public health challenges.
\bibliographystyle{splncs04}
\bibliography{samplepaper}




% LNCS Homepage, \url{http://www.springer.com/lncs}, last accessed 2023/10/25
% \end{thebibliography}
\end{document}